\documentclass[journal]{IEEEtran}
\ifCLASSINFOpdf
\else
   \usepackage[dvips]{graphicx}
\fi
\usepackage{url}

\usepackage{graphicx}
\usepackage{stfloats}
\usepackage{lineno}
\usepackage{times}

\usepackage{amsmath}
\usepackage{amssymb}
\usepackage{subcaption}
\usepackage{tabularx}
\usepackage{booktabs}
\usepackage{bm}

\usepackage{color}
\usepackage{textcomp}
\usepackage{hyperref}

\begin{document}

\title{Heatmap-based Vanishing Point boosts Lane Detection}

\author{Yin-Bo Liu, Ming Zeng, \IEEEmembership{Member, IEEE}, and Qing-Hao Meng, \IEEEmembership{Member, IEEE}
\thanks{This work is supported by the National Natural Science Foundation of China (No. 61573253), and National Key R\&D Program of China under Grant No. 2017YFC0306200.}
\thanks{The authors are with the Institute of Robotics and Autonomous Systems, Tianjin Key Laboratory of Process Measurement and Control, School of Electrical and Information Engineering, Tianjin University. Corresponding author: Ming Zeng (e-mail: zengming@tju.edu.cn) and Qing-Hao Meng (e-mail:qh\_meng@tju.edu.cn).}
}
\maketitle
\begin{abstract}
Vision-based lane detection (LD) is a key part of autonomous driving technology, and it is also a challenging problem. As one of the important constraints of scene composition, vanishing point (VP) may provide a useful clue for lane detection. In this paper, we proposed a new multi-task fusion network architecture for high-precision lane detection. Firstly, the ERFNet was used as the backbone to extract the hierarchical features of the road image. Then, the lanes were detected using image segmentation. Finally, combining the output of lane detection and the hierarchical features extracted by the backbone, the lane VP was predicted using heatmap regression. The proposed fusion strategy was tested using the public CULane dataset. The experimental results suggest that the lane detection accuracy of our method outperforms those of state-of-the-art (SOTA) methods.
\end{abstract}


\IEEEpeerreviewmaketitle

\section{Introduction}
In recent years, autonomous driving technology \cite{lee2020unsupervised} has become one of the most popular investment directions in the field of artificial intelligence. As an important part of autonomous driving, lane detection (LD) has attracted much attention from the researchers \cite{li2017deep}. At present, the performance of lane detection algorithms is acceptable for simple scenarios. However, the performance of the lane detection algorithms declines significantly for the scenes in harsh environments \cite{zhou2019deep}, such as dim light, shadow, etc.

Existing lane detection algorithms can be divided into two categories: deep learning (DL) based and traditional non-DL based. Traditional non-DL algorithms first extract hand-crafted features, then post-process these hand-crafted features, and finally obtain the estimated results of lanes. The commonly used hand-crafted features include color features \cite{color07}, line segment detection (LSD) features \cite{lsd08} and Hough transform features \cite{hough13}, etc. Traditional non-DL methods have two shortcomings: 1) The hand-crafted features are shallow features of the scenes, so their representation capabilities are limited and they are susceptible to scene noise; 2) The feature integration ability of post-processing methods is also limited. Therefore, the detection performance of the non-DL methods is not ideal. 

In recent years, deep learning technology has made a series of major breakthroughs in the field of image analysis. Therefore, researchers attempt to use DL technology to solve the challenging problem in complex scenes. For example, Neven et al. \cite{toward18} propose a semantic instance segmentation method, which can achieve end-to-end lane detection. Pan et al. \cite{scnn} optimize the extraction of spatial information in the image using the SCNN network. Hou et al. introduce self-attention distillation (SAD) \cite{hou} and Inter-Region Affinity Knowledge Distillation (IntRA-KD) \cite{hou2020inter} into the lane detection, which improves detection performance while reducing parameters. Philion \cite{philion2019fastdraw} and Liu et al. \cite{liu2020lane} introduce the style transfer network into lane detection to solve the problems of long tail and low light condition. Yoo et al. \cite{yoo2020end}  translate the lane marker detection problem into a row-wise classification task, which performs the prediction in an end-to-end manner.

As one of the important constraints of scene composition, vanishing point (VP) can also provide important clues for lane detection \cite{su2018vanishing}. For straight lanes, the VP is the intersection of lanes in the distance \cite{yoo2017a}. For the curved lanes \cite{shi2015fast}, the VP is the intersection of Lane tangents. In some non-DL algorithms, researchers attempt to use the VP as a constrain to assist lane detection. However, due to the low accuracy of VP detection, these algorithms have not been widely used. In a DL based algorithm, Lee et al. \cite{lee2017vpgnet} combine the output of the binary graph and the 4-quadrant distribution map to determine the VP of a scene, and simultaneously predict the lane. Although this algorithm improves the accuracy of lane detection, it is very difficult to integrate it with the classic CNN-based target detection architecture due to its lack of versatility and difficulty in labeling.

In this paper, we proposed a new VP-assisted lane detection method based on heatmap regression. The heatmap regression \cite{newell2016stacked} can perform pixel-level estimation of key points in the image and achieve very good results in the application of 2D human pose estimation. We also found that heatmap regression can be used to detect the VP in the road scene images.  In order to better integrate the VP detection with lane detection, we proposed a new multi-task fusion network architecture. In the experimental analysis, we systematically investigated the effectiveness of the proposed fusion strategy on public dataset.

The main contributions are as follows:

\begin{itemize}

\item We proposed a lane VP detection algorithm based on heatmap regression, which can obtain high-precision VP detection results.
\item A new multi-task fusion network architecture was proposed, which can well integrate VP detection task and lane detection task, and significantly improve the lane detection accuracy.

\end{itemize}

The remainder of this paper is organized as follows. We first introduce our proposed lane detection algorithm which includes heatmap-based lane VP detection and a multi-task fusion architecture in Section \ref{sec2}. In Section \ref{sec3}, we evaluate the performance of the proposed algorithm, followed by the conclusions in Section \ref{sec4}.

\section{Methodology}\label{sec2}

\begin{figure*}[t]
\centering
\includegraphics[scale=0.5]{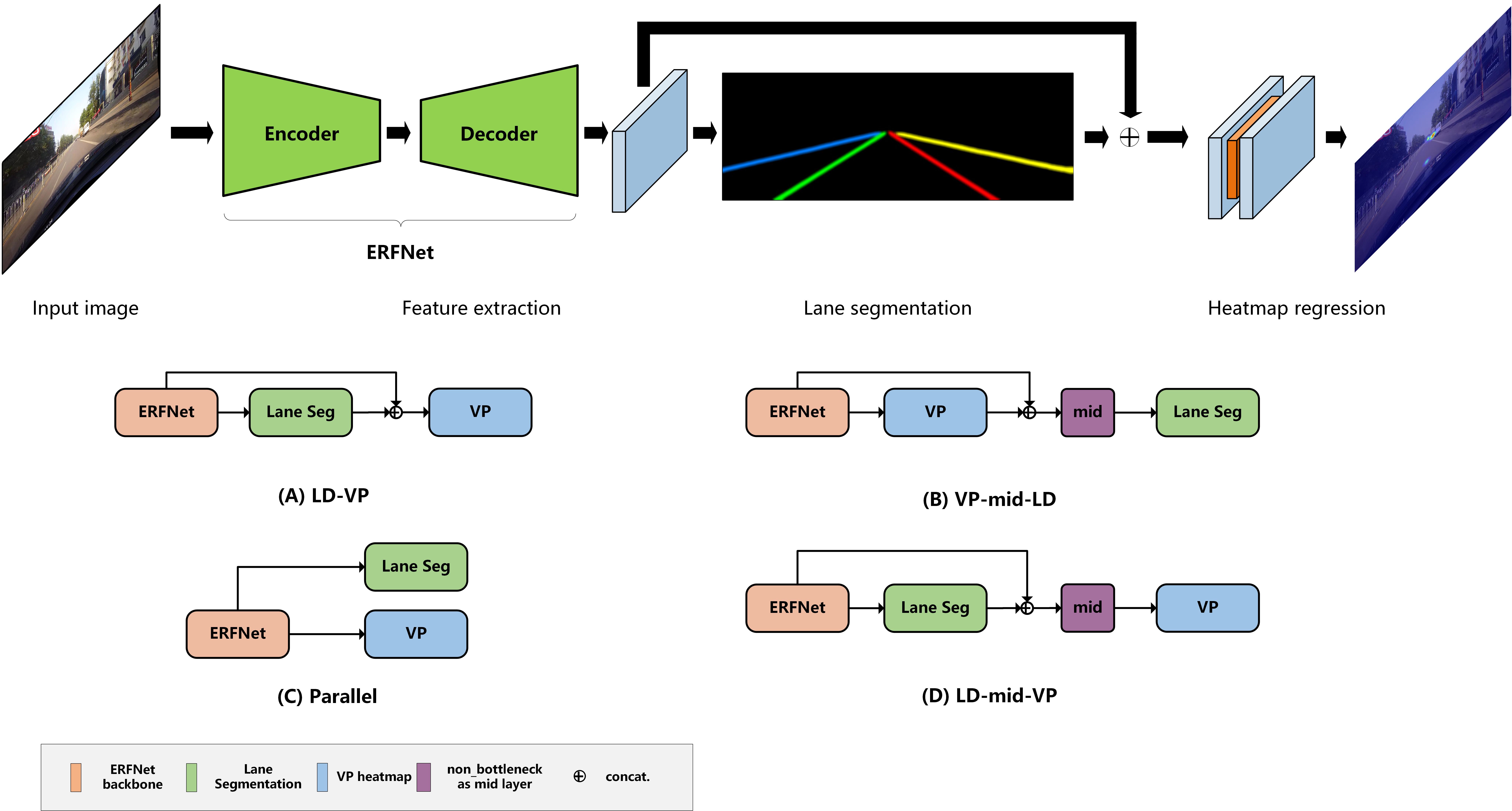}
\caption{Illustration of our proposed network architecture and four possible structures. The ERFNet is the backbone of the detection network. We finally choose the structure (D), which combines the output of lane detection and the hierarchical features extracted by the ERFNet to predict the lane VP using heatmap regression. }
\label{fig:label1}
\end{figure*}

The VP of the road provides an important clue for lane detection, but there are two difficulties in how to effectively introduce VP information into the CNN-based lane detection algorithm: 1) how to predict the VP of the road with high accuracy; 2) how to effectively integrate VP detection and lane detection. In view of the above difficulties, we proposed a new multi-task fusion network architecture. Firstly, the ERFNet \cite{romera2018erfnet:} is used as the backbone to extract the hierarchical features of the road image. Then, the lanes are detected using image segmentation. Finally, combining the output of lane detection and the hierarchical features extracted by the backbone, the lane VP is predicted using heatmap regression. This fusion strategy can not only achieve high-precision VP estimation but also solve the problem of multi-task loss function unbalance. The overall architecture of the network is shown in Figure \ref{fig:label1}.

\subsection{VP detection based on Heatmap Regression}\label{sec2.1}

The previous research results show that heatmap regression is a good keypoint detection technique, which can perform a pixel-level estimation of keypoints in the image. At present, this technique has achieved good detection results in 2D human pose estimation applications. In this research, we find that the lane VP can be regarded as a special keypoint. Therefore, CNN-based heatmap regression can be applied to predict the lane VP  end-to-end.

\subsection{Combining structure}\label{sec2.2}
There are several ways to integrate the lane detection task with the VP task: A) LD-VP structure, i.e., the result of lane detection is combined with the hierarchical features of ERFNet as the input of VP detection; B) VP-mid-LD structure, i.e. the result of VP detection is combined with the hierarchical features of ERFNet as the input of a middle layer and the output of the middle layer is used as the input of lane detection; C) parallel structure, i.e. VP detection and lane detection are independent; D) LD-mid-VP structure, i.e. the result of lane detection is combined with the hierarchical features of ERFNet as the input of a middle layer and the output of the middle layer is used as the input of VP detection. The middle layer mentioned above is a non-bottleneck block in ERFNet. After extensive testing, we find that the LD-mid-VP structure can achieve the best lane detection results compared to the other three structures. Fig \ref{fig:label1} shows the four possible structures.

\subsection{Loss Function}\label{sec2.3}
To train our complete network, we minimize the following loss function.
\begin{equation}
L_{oss}=\lambda_{vp}l_{vp}+\lambda_{lane}l_{lane},
\end{equation}

\noindent where $l_{vp} $ and $l_{lane} $ are heatmap loss and lane detection loss, respectively. we use the mean-squared error for the VP heatmap loss and cross entropy losses for lane detection. $ \lambda_{vp} $ and $\lambda_{lane}$ are the training weight of VP loss and lane detection loss, respectively. In order to balance the tasks of VP detection and lane detection, we set $\lambda_{vp}$ to 15 and $\lambda_{lane}$ to 1.

\section{Experiments}\label{sec3}
\subsection{Dataset construction}
In order to compare the performance of different lane detection algorithms, we selected the widely used CULane \cite{scnn} dataset. This dataset contains 123K images from many different challenging driving scenarios, like \textit{Dazzle light}, \textit{Crowed}, \textit{Night}, \textit{Shadow}, and so on. However, the CULane dataset does not contain the labeled VP information. Therefore, we manually annotated the VPs for the CULane dataset.

\begin{table*}[t]
\centering
\caption{Comparison of F1-measure and running time for different approaches on CULane test set. For crossroad, only FP is shown.}
\begin{tabular}{c|c|c|c|c|c|c|c|c|c|c|c}
\hline
Category     & Proportion & \textbf{\begin{tabular}[c]{@{}c@{}}ERFNet\\ -VP\end{tabular}} & ERFNet & \begin{tabular}[c]{@{}c@{}}ERFNet\\ -E2E \cite{yoo2020end}\end{tabular}  & \begin{tabular}[c]{@{}c@{}}CycleGAN\\ -ERFNet \cite{liu2020lane}\end{tabular}  & \begin{tabular}[c]{@{}c@{}}ENet\\ -SAD \cite{hou}\end{tabular} &\begin{tabular}[c]{@{}c@{}} SCNN\\ \cite{scnn} \end{tabular} & \begin{tabular}[c]{@{}c@{}}Res18\\ \cite{chen2017deeplab}\end{tabular}        & \begin{tabular}[c]{@{}c@{}}Res18\\ -VP\end{tabular}  & \begin{tabular}[c]{@{}c@{}}Res34\\ \cite{chen2017deeplab}\end{tabular}  & \begin{tabular}[c]{@{}c@{}}Res34\\ -VP\end{tabular} \\ \hline
Normal       & 27.7\%     & \textbf{91.9}                                                & 91.5   & 91.0                                                  & 91.8                                                       & 90.1                                                & 90.6  & 84.9          & 89.2   & 88.1                                                         & 90.4                                                         \\
Crowded      & 23.4\%     & 72.3                                                         & 71.6   & \textbf{73.1}                                         & 71.8                                                       & 68.8                                                & 69.7  & 63.8          & 67.9   & 67.0                                                         & 69.2                                                         \\
Night        & 20.3\%     & \textbf{69.4}                                                & 67.1   & 67.9                                                  & 69.4                                                       & 66.0                                                & 66.1  & 58.1          & 62.6   & 59.4                                                         & 63.8                                                         \\
No line      & 11.7\%     & \textbf{46.8}                                                & 45.1   & 46.6                                                  & 46.1                                                       & 41.6                                                & 46.4  & 36.3          & 41.7   & 40.7                                                         & 43.1                                                         \\
Shadow       & 2.7\%      & 74.0                                                         & 71.3   & 74.1                                                  & \textbf{76.2}                                              & 65.9                                                & 66.9  & 49.7          & 58.8   & 58.8                                                         & 62.5                                                         \\
Arrow        & 2.6\%      & 87.4                                                         & 87.2   & 85.8                                                  & \textbf{87.8}                                              & 84.0                                                & 84.1  & 75.4          & 81.6   & 80.7                                                         & 83.5                                                         \\
Dazzle light & 1.4\%      & \textbf{67.1}                                                & 66.0   & 64.5                                                  & 66.4                                                       & 60.2                                                & 58.5  & 50.4          & 59.3   & 56.8                                                         & 61.4                                                         \\
Curve        & 1.2\%      & 66.4                                                         & 66.3   & \textbf{71.9}                                         & 67.1                                                       & 65.7                                                & 64.4  & 53.2          & 60.8   & 58.8                                                         & 64.7                                                         \\
Crossroad    & 9.0\%      & 2292                                                          & 2199   & 2022                                                  & 2346                                                       & 1998                                                & \textbf{1990}  & 2452 & 2919   & 2667                                                         & 2141                                                         \\
Total        & -          & \textbf{74.2}                                                & 73.1   & 74.0                                                  & 73.9                                                       & 70.8                                                & 71.6  & 65.1          & 69.1   & 67.8                                                         & 70.9                                                         \\ \hline
Runtime (ms)  & -          & 10.4                                                          & 8.9    & -                                                     & -                                                          & 11.7                                                & 116.5 & 22.5          & 23.7   & 24.1                                                         & 25.6                                                         \\
Parameter (M) & -          & 2.492                                                         & 2.488  & -                                                     & -                                                          & 0.98                                                & 20.72 & 16.062        & 16.069 & 25.702                                                       & 25.709                                                       \\ \hline
\end{tabular}
\label{tab1}
\end{table*}

\subsection{Metrics}

We used the method proposed in \cite{scnn} to quantitatively evaluate the lane detection performance of various algorithms. We treated each lane marking as a line with 30-pixel width and computed the intersection-over-union (IoU) between labels and predictions. Predictions whose IoUs were larger than a threshold were considered as true positives (TP). Here, the threshold was set to 0.5. Then, we used F1 measure as the evaluation metric, which is defined as: $F1 = \frac{2 \times Precision \times Recall}{Precision + Recall}$, where $Precision = \frac{TP}{TP + FP}$, and $Recall = \frac{TP}{TP + FN} $.

\subsection{Implementation Details}

We implemented our method in Python using Pytorch 1.3 and CUDA 10 and ran it on an i7-8700K@3.7GHz with NVIDIA RTX 2080 Ti. We used the CULane's training dataset as the training dataset, which contains 88,880 images, and the CULane's test dataset as the test dataset (34,680 images). All input images were reshaped to 976 $ \times $ 351 for training. We applied a Gaussian kernel with the same standard deviation (std = 7 by default) to all these ground truth heatmaps. We used stochastic gradient descent (SGD) for optimization and started with a learning rate of 0.001 for the network. We divided the learning rate by 10 every 5 epochs, with a momentum of 0.9. We also adopted data augmentation with random flip and image rotations.

\subsection{Comparisons}

\begin{figure*}[htbp]
\centering
\includegraphics[scale=0.38]{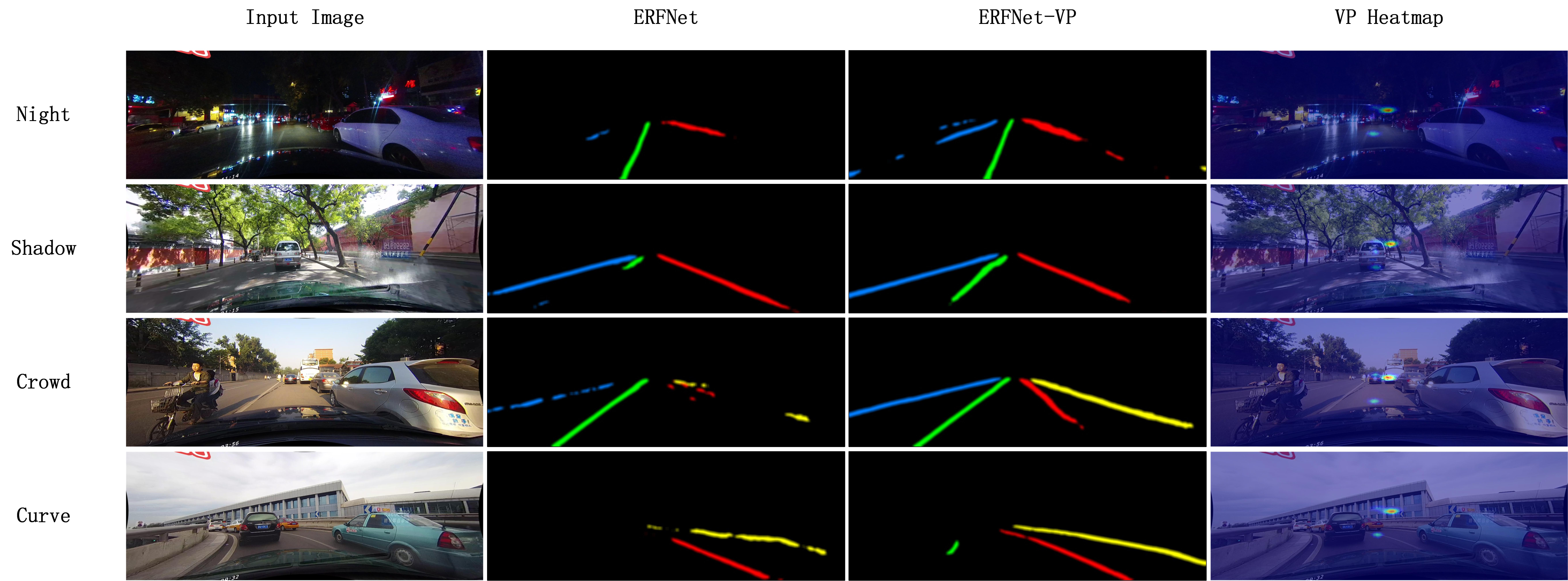}
\caption{Some sample images from CULane dataset. From left to right are the input images, results of ERFNet, the results of ERFNet-VP and the predicted VPs.}
\label{fig:label3}
\end{figure*}

Table \ref{tab1} shows the test results of the proposed algorithm on the CULane testset. From left to right are the results of our method, ERFNet, ERFNet-E2E, CycleGAN-ERFNet, ENet-SAD, SCNN, ResNet18, ResNet34, ResNet18-VP and ResNet34-VP. It is obvious that the proposed algorithm is superior to the SOTA in most groups, especially in \textit{Normal}, \textit{Night}, and \textit{Dazzle light} groups. The overall F1 measurement accuracy has been improved from 73.1 to 74.2. It is easy to see that using the lane VP information can improve the detection accuracy in dazzle light and night conditions, and multitask learning can greatly improve the overall detection performance of the network.

Fig. \ref{fig:label3} illustrates the different performances between our method and the ERFNet in \textit{Night}, \textit{Shadow}, \textit{Crowded}, and \textit{Curve}. The positions of the lane VP are marked by heatmap. It can be seen that the probability maps generated by our method are more accurate than those of the ERFNet.

\subsection{Ablation Study}

Backbone selection, lane VP detection sub-network and the multi-task fusion architecture are three key factors affecting the final lane detection results. Therefore, we made ablation study to quantitatively analyze the influence of the key factors on the performance of the lane detection.

\subsubsection{Backbone}
We systematically tested the effects of different backbones, i.e., ResNet18 (Res18), ResNet34 (Res34) and ERFNet, on accuracy and detection speed of the model. As shown in Table \ref{tab1}, the accuracy of VP assisted lane detection is better than that of non-VP assisted prediction when choosing any kind of backbone. More Specifically, for VP-based networks using Res18, Res34, and ERFNet as backbones, the results of F1-measure have increased by 4.0\%, 3.1\%, and 1.1\%, respectively, compared with non-VP counterparts. At the same time, we also note that adding the part of VP detection has little effect on the running time of lane detection model (about 1ms additional calculations).

\subsubsection{Lane VP Detection}
In order to quantitatively evaluate the lane VP detection performance of our algorithms, we used the normalized Euclidean distance proposed in \cite{moghadam2011fast} to measure the estimated errors between the detected lane VP and the manually labeled ground truth. The standardized Euclidean distance is defined as:

\begin{equation}
NormDist=\frac{\lVert P_{g}-P_{v} \rVert}{Diag(I)},
\end{equation}
where $P_{g}$ and $P_{v}$ denote the ground truth of the lane VP and the estimated lane VP, respectively. $Diag(I)$ is the length of the diagonal of the input image. The closer the $NormDist$ is to 0, the closer the estimated lane VP is to the ground truth. The $NormDist$ greater than 0.1 is set to 0.1, which is considered to be a failure of the corresponding method. 

\begin{figure}[]
\centering
\includegraphics[scale=0.5]{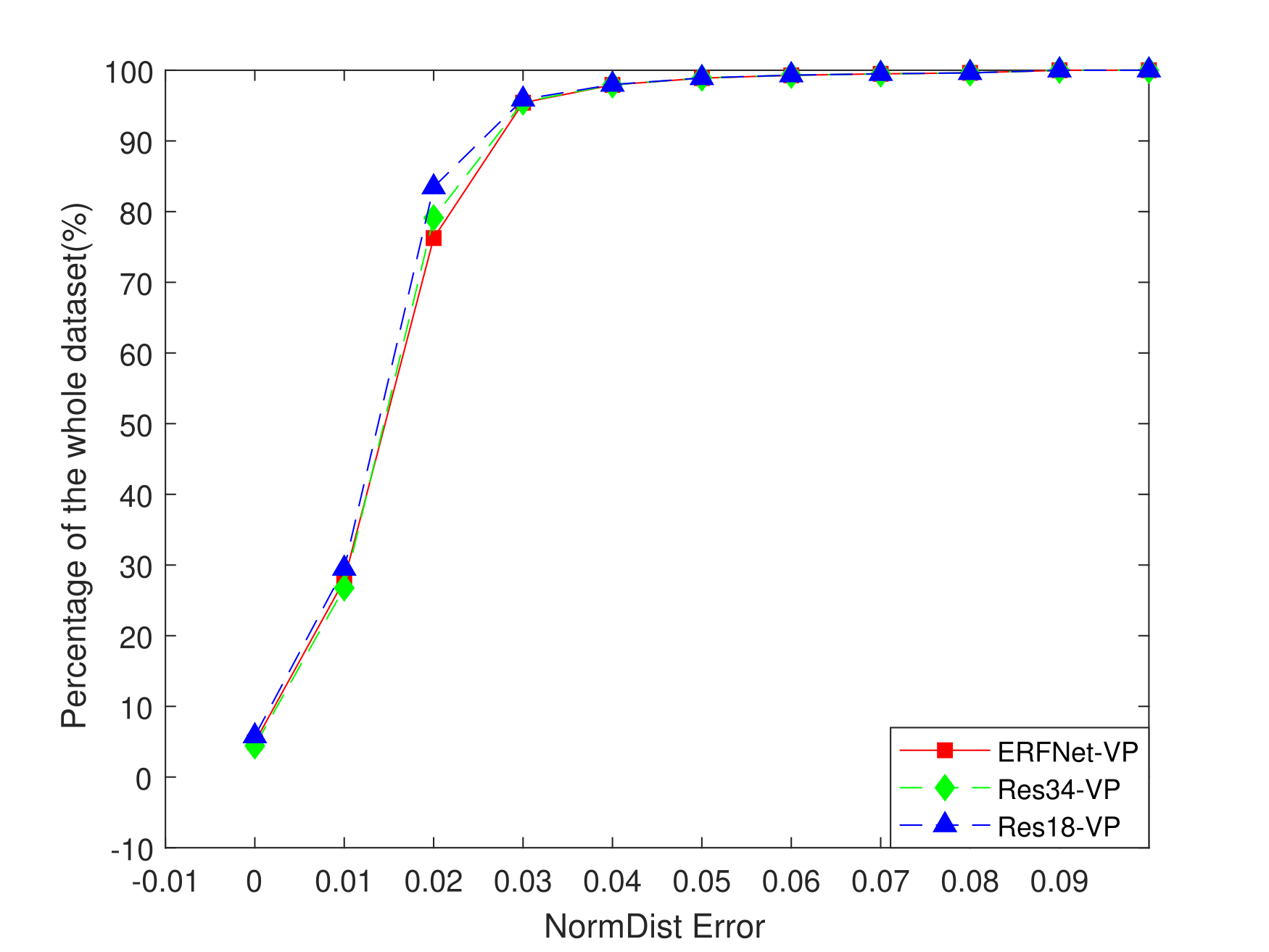}
\caption{Accumulated error distribution of our lane VP detection on the CULane dataset. On the x-axis, 0 stands for $NormDist$ in [0, 0.01), 0.01  stands for $NormDist$ in [0.01, 0.02)..., and 0.1 represents $NormDist$ in [0.1, 1].}
\label{fig:label4}
\end{figure}

The Fig. \ref{fig:label4} shows the results using different backbones on the CULane dataset. For the models whose backbones are ERFNet, ResNet18, and ResNet34, the corresponding proportions of the small detection errors (NormDist \textless 0.01) are 4.93\%, 4.41\% and 5.79\%, and the proportion of large detection errors (NormDist \textgreater 0.05) are 2.07\%, 2.02\% and 2.05\%, respectively. In addition, the mean errors of NormDist are 0.024859, 0.024984, and 0.024049, respectively. These results indicate that the selection of backbone has little effect on the performance of the detection model.

\subsubsection{Combining Structure}

\begin{table}[]
\centering
\caption{Comparison results of F1-measure and running time for different combining structures on CULane test set.}
\begin{tabular}{c|c|c|c|c}
\hline
Category     & LD-VP & VP-mid-LD      & Parallel & LD-mid-VP      \\ \hline
Normal       & 91.5 & 91.6          & 91.6    & \textbf{91.9} \\
Crowded      & 71.6  & \textbf{72.4} & 72.2    & 72.3          \\
Night        & 69.0 & \textbf{69.6}  & 68.4    & 69.4          \\
No line      & 45.8 & 45.8          & 46.6    & \textbf{46.8} \\
Shadow       & 74.0    & \textbf{74.8} & 71.9    & 74.0          \\
Arrow        & 86.4 & 86.7          & 85.4    & \textbf{87.4} \\
Dazzle light & 65.9  & 64.4           & 65.3    & \textbf{67.1} \\
Curve        & 65.3  & 66.0          & 65.8    & \textbf{66.4} \\
Crossroad    & 2248  & \textbf{1997}  & 2321     & 2292           \\
Total        & 73.6 & 74.1          & 73.7    & \textbf{74.2}
\end{tabular}
\label{tab2}
\end{table}
As mentioned in Section \ref{sec2.2}, there are four alternative structures. We quantitatively evaluate the impact of the selection of different structures on the lane detection performance. The results are shown in Table \ref{tab2}. The corresponding results of F1-measure for the LD-VP, VP-mid-LD, Parallel, and LD-mid-VP are 73.6, 74.1, 73.7, and 74.2, respectively. The results of structure (B) is better than that of structure (D) in the four categories of \textit{Crowded}, \textit{Night}, \textit{Shadow} and \textit{Crossroad}, but the overall result is not as good as that of structure (D). Therefore, we selected the structure (D) as the multi-task fusion network.

\section{Conclusions}\label{sec4}
Vanishing point (VP) is an important clue for lane detection. In this paper, we proposed a new multi-task fusion network architecture in which the VP information extracted by heatmap regression can benefit for the lane detection. We selected the LD-mid-VP structure with better performance as the fusion structure from four possible structures. Experimental results show that our proposed method has the advantages of high accuracy and robustness under the working conditions such as shadow, night, and curve. It would be interesting to extend this idea to other tasks that demand VP assistance, such as image retrieval and pose estimation.

\bibliography{main}

\end{document}